\documentclass[final]{cvpr}

\usepackage{times}
\usepackage{epsfig}
\usepackage{graphicx}
\usepackage{amsmath}
\usepackage{amssymb}
\usepackage{multirow}
\usepackage{makecell}
\usepackage{subcaption}
\usepackage{float}

\usepackage[pagebackref=true,breaklinks=true,colorlinks,bookmarks=false]{hyperref}
\def\cvprPaperID{10687} 
\def\confYear{CVPR 2021}

\begin{document}

\title{Learning Better Visual Dialog Agents with Pretrained Visual-Linguistic Representation}

\author{

Tao Tu\textsuperscript{1},
Qing Ping\textsuperscript{2,}\thanks{Corresponding author} , 
Govind Thattai\textsuperscript{2},
Gokhan Tur\textsuperscript{2},
Prem Natarajan\textsuperscript{2}
\\

\textsuperscript{1}National Taiwan University \\ \textsuperscript{2}Amazon Alexa AI\\
{\tt\small ttaoREtw@gmail.com,}
{\tt\small\{pingqing,thattg,gokhatur,premknat\}@amazon.com}}

\maketitle

\begin{abstract}
\vspace{-3pt}

GuessWhat?! is a visual dialog guessing game which incorporates a Questioner agent that generates a sequence of questions, while an Oracle agent answers the respective questions about a target object in an image. Based on this dialog history between the Questioner and the Oracle, a Guesser agent makes a final guess of the target object. 
While previous work has focused on dialogue policy optimization and visual-linguistic information fusion, most work learns the vision-linguistic encoding for the three agents solely on the GuessWhat?! dataset without shared and prior knowledge of vision-linguistic representation. To bridge these gaps, this paper proposes new Oracle, Guesser and Questioner models that take advantage of a pretrained vision-linguistic model, VilBERT.  
For Oracle model, we introduce a two-way background/target fusion mechanism to understand both intra and inter-object questions. For Guesser model, we introduce a state-estimator that best utilizes VilBERT's strength in single-turn referring expression comprehension. For the Questioner, we share the state-estimator from pretrained Guesser with Questioner to guide the question generator. Experimental results show that our proposed models outperform state-of-the-art models significantly by 7\%, 10\%, 12\% for Oracle, Guesser and End-to-End Questioner  respectively.

\end{abstract}

\section{Introduction}
Multi-modal dialog tasks have gained increasing popularity in recent years such as \textit{GuessWhat?!} \cite{de2017guesswhat}, \textit{GuessWhich?!} \cite{chattopadhyay2017evaluating}, VisDial \cite{das2017visual}, VDQG \cite{li2017learning}, vision-and-language navigation R2R \cite{anderson2018vision}, ImageChat \cite{shuster2020image}, Alfred \cite{shridhar2020alfred} and so on. Multi-modal dialog tasks are challenging as the models need to perform high-level image understanding and visual grounding, and such visual grounding should be properly combined with understanding and tracking of multi-turn dialogues in the meanwhile.

\begin{figure}[t]
\centerline{\includegraphics[width=0.7\linewidth]{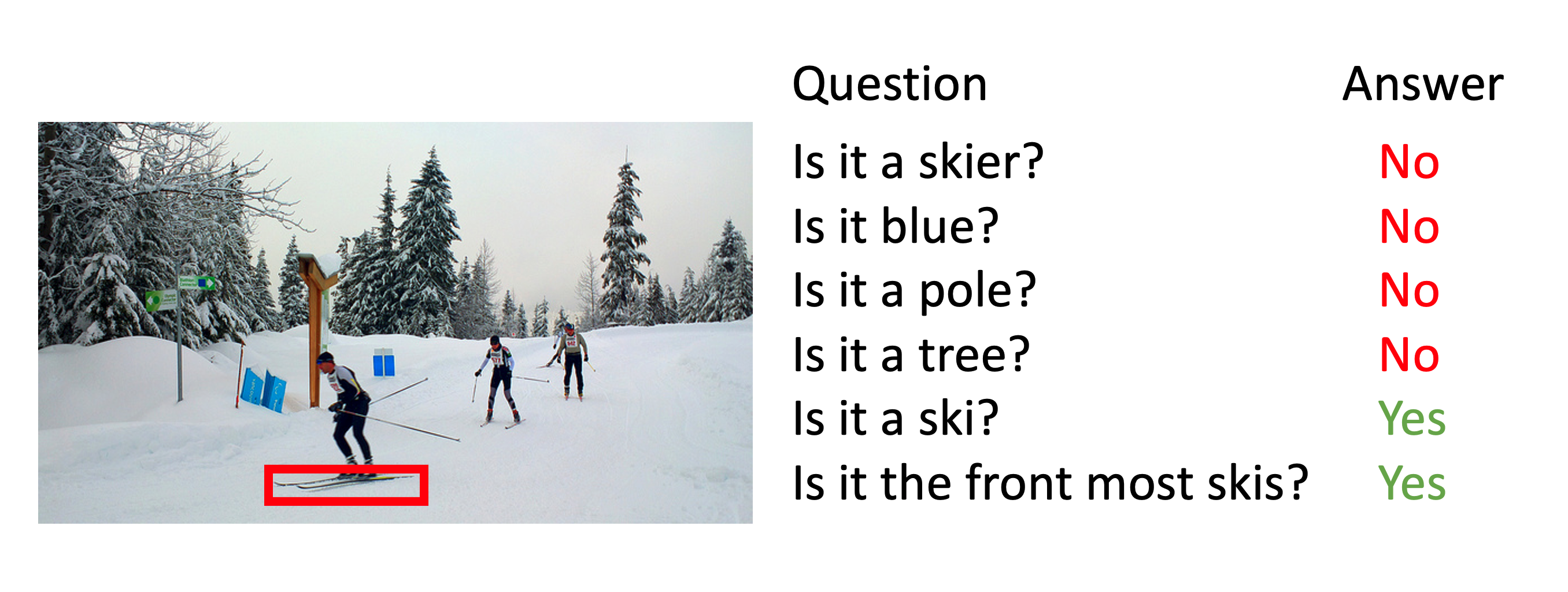}}

\caption{
Example of the \textit{GuessWhat?!} dataset
}

\label{fig:guessswhat_example}
\end{figure}

The \textit{GuessWhat?!} dataset is a challenging dataset for a two-player game, where one player will ask a sequence of binary questions and make a final guess for an object in an image designated by another player. 
The first player performs two sub-tasks, namely as a Questioner to ask questions, and as a Guesser to make the final guess. The second player serves as the Oracle to give Yes/No answer to first player's questions.  An example of the \textit{GuessWhat?!} game can be seen in Figure-\ref{fig:guessswhat_example}. The \textit{GuessWhat?!} game is a good test-bed for such multi-modal tasks such as VQA, referring expression comprehension and generation, and it is also organized in a multi-turn multi-agent dialog. This paper focuses on the three agents for the two players in the \textit{GuessWhat?!} dataset, namely the Oracle model, Guesser model and the Questioner model. 

The Oracle task can be considered as an object-aware Visual Question Answering task (VQA), where the inputs are an image, a question, and a pre-defined target object, and the output is an answer of Yes/No/NA depending on whether the question matches the target object. The baseline Oracle model \cite{de2017guesswhat} encodes the target object with only category and spatial information but no visual information, which may be insufficient for answering more complex questions about color, shape, relation, actions of an object. To bridge this gap, we introduce VilBERT-Oracle, which takes advantage of the VilBERT model's ability to achieve state-of-the-art performance on VQA tasks \cite{lu2019vilbert, lu202012}. We also introduce a two-way background/target fusion mechanism on top of the VilBERT encoder to learn how to predict correct binary answers with respective to a target object. 

The Guesser model can be considered as a special case of referring expression comprehension problem.
Given an image and an entire dialog history of questions (referring expression) and its corresponding answers, the Guesser model has to look at the entire dialog and make a final guess. One intuitive solution is to simply concatenate the entire dialog and feed them to the model \cite{deng2018visual,strubend,shukla2019should,lu202012}. However, this might be inadequate if the dialog history is not properly dissected in a way to promote/demote objects according to question and answer in each turn. Recent work \cite{pang2020visual} introduces object state tracking mechanism, where the belief of all objects is dynamically updated after each turn. Another issue is that almost all existing Guesser models learn the vision-linguistic associations between object and question from scratch on the \textit{GuessWhat?!} dataset, which may be sparse in coverage of referring expressions for new objects. To bridge this gap, we propose VilBERT-Guesser, which is built on top of VilBERT's strength in single-turn referring expression comprehension, and introduces the object state tracking mechanism into VilBERT encoder to learn to update the belief of object states throughout the dialog. To our best knowledge, this is the first work that brings dialog state tracking to large-scale pre-trained vision-linguistic model, which is meant to work only on single-turn text descriptions.  

The Questioner model can be considered as a special case of referring expression generation problem.
Previous works for Questioner model intuitively encode the image feature and dialog history information to a fused representation, and utilize a language decoder to generate the question \cite{strubend, shekhar2018beyond,zhao2018improving,zhang2017asking,abbasnejad2018active,abbasnejad2019s}. The multi-modal fusion modules are mostly learned from scratch on the \textit{GuessWhat?!} dataset, which may be insufficient for similar reasons as the Guesser models. Moreover, the encoding of the dialog history as a whole, poses challenges for language generator which tends to forget long-term history and generates repeated questions. Recent work introduces state-tracking to Questioner, which dynamically feeds the updated beliefs over objects into Questioner, so that the language generator could generate more targeted questions in each turn\cite{pang2020visual}. Inspired by this work, we introduce object state estimation mechanism to our VilBERT-Questioner. Moreover, once the VilBERT-Guesser is trained, we load its weights to the state-estimator of VilBERT-Questioner, so that the later could take advantage of the Guesser's ability to make reliable predictions for estimating object states. 

Our major contribution of the paper are as follows. First, we propose novel Oracle, Guesser and Questioner models that are built on top of a state-of-the-art vision-linguistic pre-trained model. The proposed models outperform existing state-of-the-art models with significant margins. Second, we propose a unified framework for Guesser and Questioner so that Questioner can take advantage of the robust state-estimator learned from VilBERT-Guesser. Third, we conduct thorough ablation-study and analysis and find that a shared vision-linguistic representation cross the three agents may be beneficial for mutual-understanding and end-game success. Our code is made publicly available. \footnote{\url{https://github.com/amazon-research/read-up}}
\section{Related Work}
\subsection{Oracle}

The original work for \textit{GuessWhat?!} proposes a baseline Oracle \cite{de2017guesswhat} that concatenates question encoding, along with the spatial and category information of the target object, and feeds it into a MLP layer to predict the final answer. However, the baseline Oracle may be insufficient to deal with more challenging visual questions such as colors, shapes, relations, actions and so on, without visual information encoding. The Oracle task can be considered as a special case of Visual Question Answering (VQA) problem with an extra input of object identifier. Methods proposed in \cite{yang2016stacked,kim2018bilinear,jiang2018pythia,yu2019deep} achieve competitive performance on VQA tasks. However these models cannot be readily used in this task, unless they are adapted to the extra input of object identifier.


\subsection{Guesser}
The Guesser model plays an important role in the \textit{GuessWhat?!} game, which should perform both referring expression comprehension on the dialog to describe the visual objects, and perform multi-turn dialog reasoning. Earlier work proposes Guesser models that fuse the encoding of entire dialog with the object category and spatial embedding \cite{de2017guesswhat,strubend} to predict the target object. Later work in \cite{shukla2019should,lu202012,deng2018visual} adopted similar approaches and treated entire dialog history as a whole. This might be problematic for two reasons. First, reasoning over such multi-turn dialog is challenging without turn-by-turn explicit dialog state tracking. Second, the lack of turn-level visual grounding can also confuse the Guesser model as to which object the question is referring to in each turn. On the multi-modal representation, original Guesser model encodes no visual information. Some approaches have used image features such as VGG features \cite{simonyan2014very} and Faster-RCNN features\cite{yang2019making,lu2019vilbert, pang2020guessing} into Guesser models, which have shown improvement in accuracy. Recent work in \cite{pang2020guessing,yang2019making} proposes to break down the dialog into turn-level question/answer, and update the final guess with soft state tracking \cite{pang2020guessing}, which shows good performance gains.


\subsection{Questioner}
Questioner plays a key role in the GuessWhat game, since it has to both ask visually meaningful questions and guide the dialog towards goal-oriented end-game success rate. \cite{de2017guesswhat} proposed the first Questioner model with an encoder-decoder structure where dialog history is encoded with the hierarchical recurrent encoder decoder (HRED) \cite{serban2015hierarchical} and conditioned on the image which is encoded as fixed-length VGG features \cite{simonyan2014very}. Later work have introduced a shared dialog state encoder for both Guesser and Questioner, where the visual encoder is based on ResNet \cite{he2016deep} and an LSTM-based language encoder \cite{hochreiter1997long}. More recent work in \cite{pang2020visual} has incorporated turn-level object state tracking into Questioner, and has shown some improvement in a supervised learning setting. All the approaches mentioned above learn visual grounding and object state-tracking from scratch on the \textit{GuessWhat?!} dataset, which may be insufficient to generalize to new objects/games due to the sparse semantic coverage of objects.

While this paper focuses on using supervised methods for the three models, its worth mentioning there are methods as in \cite{strubend,zhang2018goal,abbasnejad2019s,zhao2018improving,pang2020visual} that use Reinforcement Learning (RL) approaches to learn Questioner/Guesser model with different variants of end-game success reward. 


\section{Vision-Linguistic Pretrained Model: Vilbert}

Before introducing our VilBERT-based models, we briefly review the model structure of VilBERT \cite{lu2019vilbert}. VilBERT is a model for learning task-agnostic joint representation of image content and natural language. Similar as the BERT architecture, VilBERT processes both visual and textual inputs in separate streams then interacts them through co-attention transformer layers \cite{lu2019vilbert}. Given an image $I$ represented as a set of object/region features $o_1, o_2, ..., o_M$ and a text input $w_1, w_2, ..., w_L$, the VilBERT model outputs final representations $h_{o1}, h_{o2},..., h_{oM}$ for vision information and $h_{w1}, h_{w2},..., h_{wL}$ for text information. For more details of VilBERT, please refer to the original work \cite{lu2019vilbert}. There are also concurrent work such as VL-Bert \cite{su2019vl}, Lxmert \cite{tan2019lxmert}, Oscar \cite{li2020oscar}, UNITER \cite{chen2019uniter} and so on.

\section{Proposed Method}
In this section, we describe the three models built upon VilBERT, namely VilBERT-Oracle, VilBERT-Guesser and VilBERT-Questioner.
  \subsection{VilBERT-Oracle Model}
The Oracle task can be considered as an object-aware Visual Question Answering task (VQA), where the inputs are an image, a question, and a pre-defined target object, and the output is an answer of (Yes, No, NA) depending on whether the question matches the target object.  

\begin{figure}[ht]
\centerline{\includegraphics[width=0.7\linewidth]{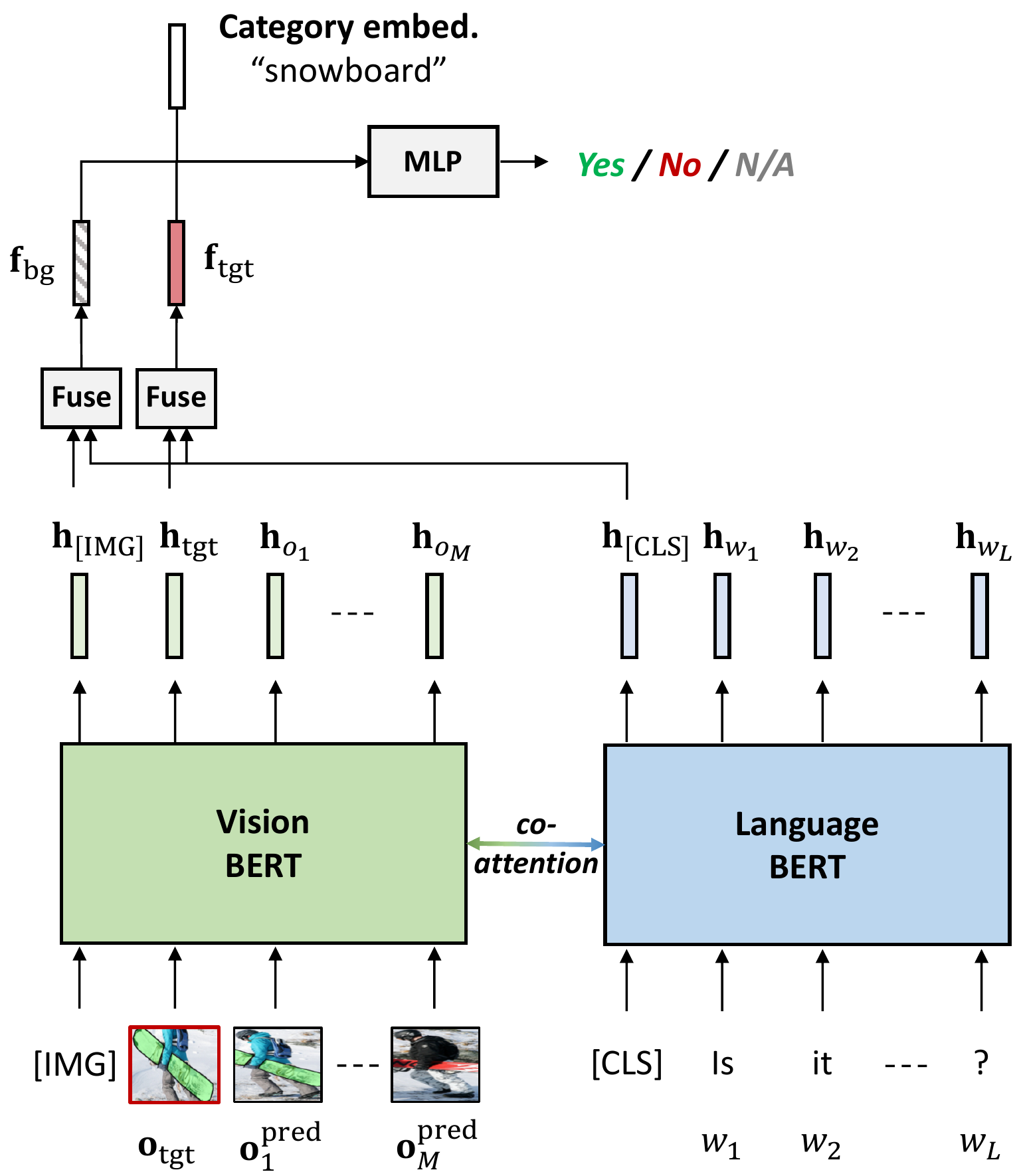}}

\caption{
Illustration of Oracle-Vilbert model.
}

\label{fig:oracle_vilbert}
\end{figure}

The Oracle model structure is illustrated in Figure-\ref{fig:oracle_vilbert}. The model is composed a multi-modal encoder (VilBERT) and a background/target fusion predictor. The multi-modal encoder includes language encoding for a question $\mathbf{q} = \{\texttt{[CLS]}, \mathbf{w}_1, \mathbf{w}_2, ..., \mathbf{w}_L\} $ and vision encoding, which in turn involves both target object encoding $\mathbf{o}_\text{tgt}$  and image/all objects encoding $\mathbf{O}_I = \{\texttt{[IMG]},  \mathbf{o}_\text{tgt}, \mathbf{o}^{\text{pred}}_1, \mathbf{o}^{\text{pred}}_2,..., \mathbf{o}^{\text{pred}}_M\}$. Features of $\mathbf{q}$ are word embeddings pretrained by Vilbert model. Features  $\mathbf{o}_\text{tgt}, \mathbf{o}^{\text{pred}}_1, \mathbf{o}^{\text{pred}}_2,..., \mathbf{o}^{\text{pred}}_M$ are visual features of target object and all $M$ regions/objects predicted by the object detection model such as Faster-RCNN \cite{ren2015faster}. The input features are then fed into Vilbert to obtain final hidden states for visual information $\mathbf{H}_O = \{\mathbf{h}_{[IMG]},  \mathbf{h}_\text{tgt}, \mathbf{h}_{o_1},..., \mathbf{h}_{o_M}\}$ and text information $\mathbf{H}_q  =\{\mathbf{h}_{[CLS]},  \mathbf{h}_{w_1},..., \mathbf{h}_{w_L}\}$.


For our two-way background/target fusion, we fuse the background image hidden states $\mathbf{h}_{\texttt{[IMG]}}$ and language output $\mathbf{h}_{\texttt{[CLS]}}$, and target object hidden states $\mathbf{h}_{\texttt{tgt}}$ and language output $\mathbf{h}_{\texttt{[CLS]}}$ respectively by taking element-wise multiplication between each pair, and concatenate them with the target object category embedding as fusion result: $\mathbf{x}_{fusion} =  (\mathbf{h}_{\texttt{[IMG]}} \odot \mathbf{h}_{\texttt{[CLS]}} ) \oplus (\mathbf{h}_{\texttt{tgt}} \odot \mathbf{h}_{\texttt{[CLS]}} ) \oplus \mathbf{c}_{cat}$. The final fusion vector $\mathbf{x}_{fusion}$ is fed into multi-layer perceptron followed by softmax to predict the answer $p_{i}$. Finally, the loss of VilBERT-Oracle is defined with cross-entropy loss on the three answer classes.

\begin{equation}
L_{VilBERT-Oracle} = - \sum_{i=1}^{N}  \sum_{j=1}^{K}    y_{i,j} \cdot log (p_{i,j})
\end{equation}

Where $N$ is total number of questions, and $K$ is the size of answer classes. 
Our intuition is that if the question matches the target object, then the fusion of $\mathbf{h}_{\texttt{tgt}}$  and $\mathbf{h}_{\texttt{[CLS]}}$ would give stronger signals. The fusion between $\mathbf{h}_{\texttt{[IMG]}}$ and $\mathbf{h}_{\texttt{[CLS]}}$ is expected to learn to understand object relations that goes beyond what $\mathbf{h}_{\texttt{tgt}}$ can represent.

  \subsection{Vilbert-Guesser Model}
The Guesser model can be considered as a special case of referring expression comprehension problem. Given an image $I$ with a set of regions/objects $\{ \mathbf{o}_1, ..., \mathbf{o}_N\} $ , and an entire dialog history of question (referring expression) and corresponding answers $\{ (q_1, a_1), ..., (q_T, a_T)\} $, a Guesser model predicts the likelihood of all objects in the image to be the target object.

\begin{figure}[t]
\centerline{\includegraphics[width=0.7\linewidth]{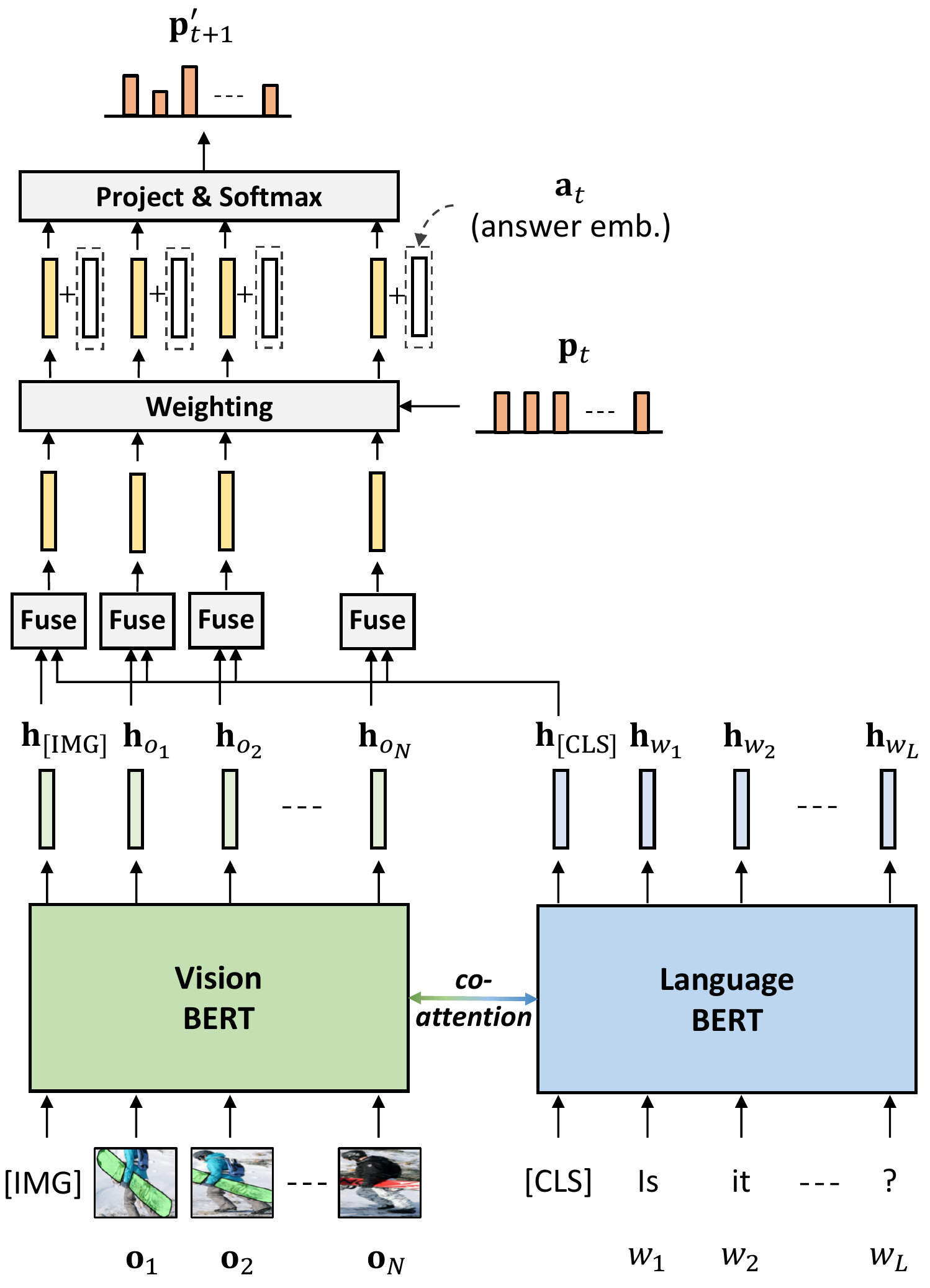}}

\caption{
Illustration of Guesser-Vilbert model.
}

\label{fig:guesser_vilbert}
\end{figure}

The model structure of VilBERT-Guesser is illustrated in Figure-\ref{fig:guesser_vilbert}. The model is composed of a multi-modal encoder (VilBERT), a global image/text fusion layer, a state-weighting layer, and answer-updating layer.  Specifically, during each turn, we first feed the visual and language features  ($\{ \mathbf{o}_1, ..., \mathbf{o}_N\} $, $(q_t)$ ) into the model, where $(q_t)$ are the word embeddings of current-turn question. After the VilBERT layers we obtain the final visual hidden states $\{ \mathbf{h}_{\texttt{<IMG>}}, \mathbf{h}_{o_1}, ..., \mathbf{h}_{o_N}\} $ and  take element-wise multiplication with the sentence-level VilBERT language output $\mathbf{h}_\texttt{<CLS>}$ for each visual state to get fused visual output $\mathbf{f}_{o_i}$: $\mathbf{f}_{o_i} = \mathbf{h}_{o_i} \odot \mathbf{h}_\texttt{<CLS>}$. Our intuition is that the fused visual output of the object $\mathbf{f}_{o_i}$ that matches the question description encoded by $\mathbf{h}_\texttt{<CLS>}$ will have stronger signals compared to irrelevant objects.  Next, the fused output for each object is weighted by previous-turn object state belief $\mathbf{p}_t$ to derive $\mathbf{f'}_{o_i}$: $\label{eq:reweight}
\mathbf{f'}_{o_i} = \mathbf{f}_{o_i} \odot p_{t_i}$, where $p_{t_i}$ is the belief of the $ith$ object in previous turn-$t$. 
Next we further update the weighted output of each object by adding the answer embedding of this turn to it: $\mathbf{v}_{o_i} = \mathbf{f'}_{o_i} + \mathbf{a}_t$. Now the final visual output of each object $\mathbf{v}_{o_i}$ should ideally satisfy all the following: (1) object(s) matching the current-turn question should have stronger signals; (2) if the objects have higher likelihood indicated in previous turn, that belief should carry over to the current turn; (3) if the answer is positive/negative, then the belief should be updated to reflect the increased/decreased belief of certain objects. The fundamental basis of all of the above is the robust referring expression comprehension that Vilbert has been pre-trained for. 

Eventually the final visual output of each object  $\mathbf{v}_{o_i}$ is fed into MLP layer followed by softmax to derive the updated state belief for this turn $\mathbf{p}_{t+1}$, which will be used to re-weight $\mathbf{f}_{o_i} $ for next-turn. We also accumulate the belief states cross-turns for faster convergence: $\mathbf{p}_{t+1} = \alpha \cdot \mathbf{p}_{t+1}' + (1-\alpha) \cdot \mathbf{p}_{t}$, where $\alpha\in[0,1]$ is the state accumulation coefficient. 
Finally, at the final turn $T$, Guesser-VilBERT makes a guess by picking the object with highest probability. Therefore the loss of VilBERT-Guesser can be defined as cross-entropy loss over all objects in an image:

\begin{equation}
L_{VilBERT-Guesser} = - \sum_{i=1}^{|D|}  \sum_{j=1}^{M}    y_{i,j} \cdot log (p_{i,j})
\end{equation}

\begin{figure*}[t]
\centerline{\includegraphics[width=\linewidth]{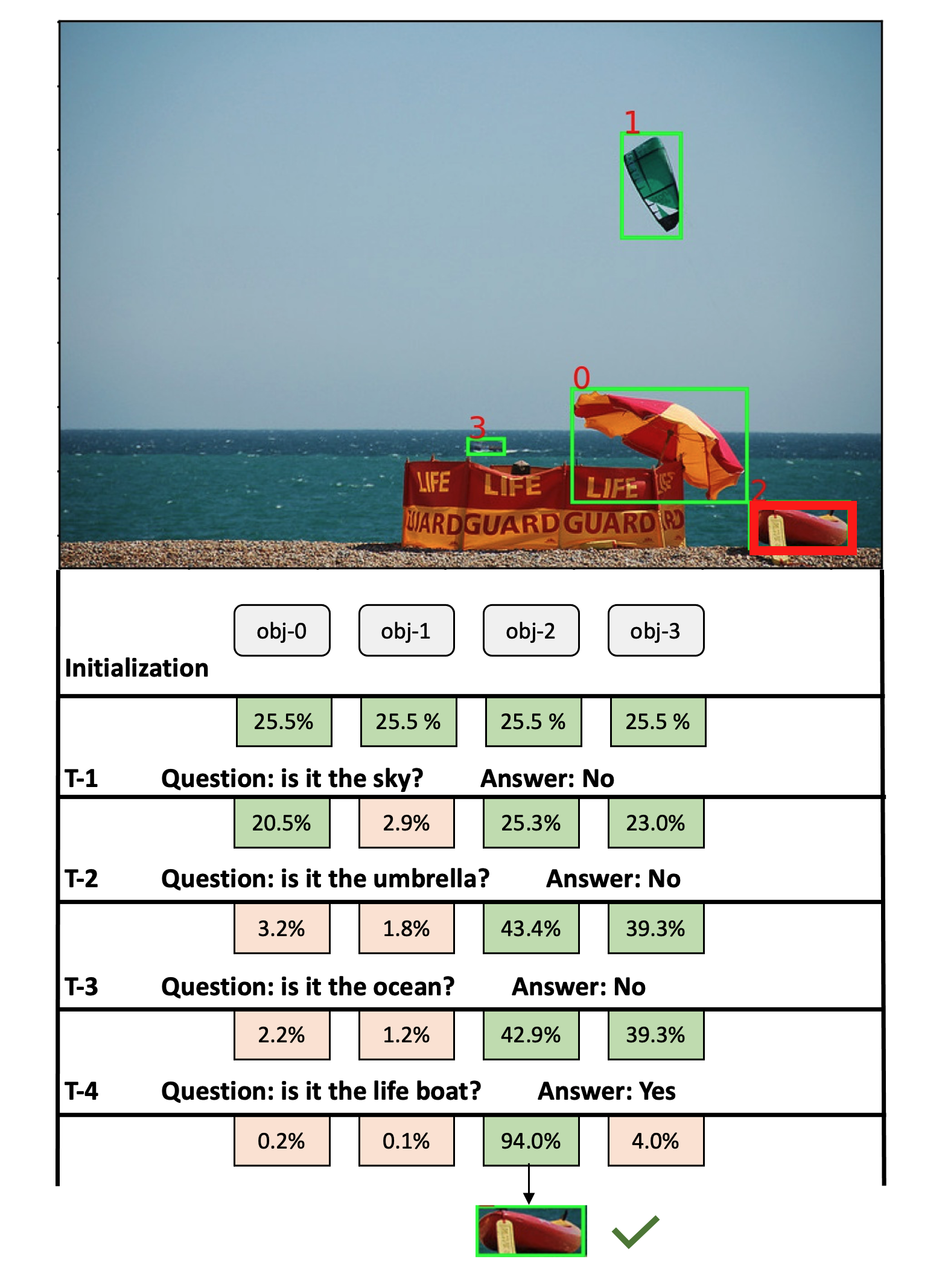}}
\caption{
Example-2 of Vilbert-Guesser State Estimator (Successful Case on Ground-Truth Dialogs)
}
\label{fig:guesser_example_2}
\end{figure*}

\begin{figure*}[t]
\centerline{\includegraphics[width=\linewidth]{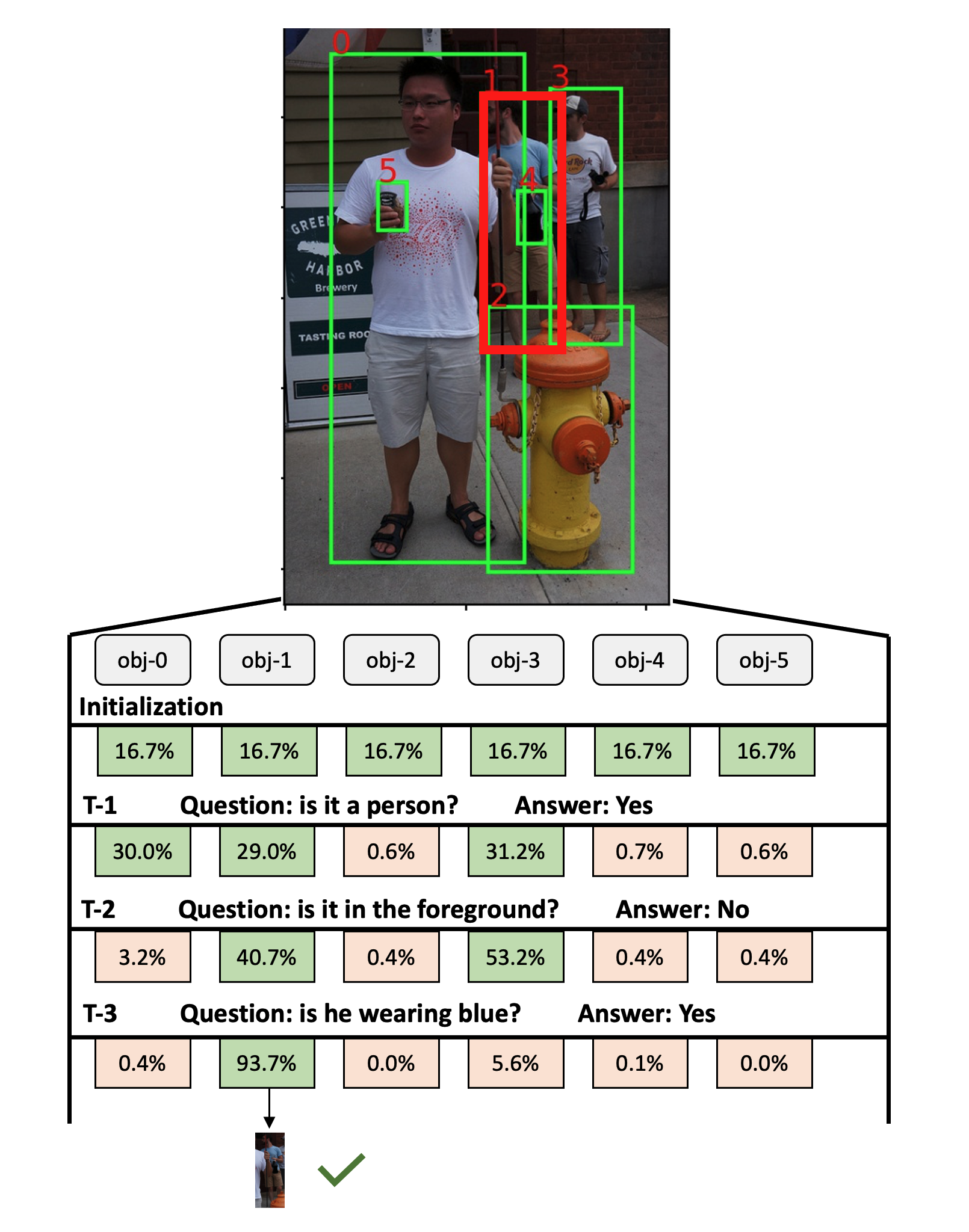}}
\caption{
Example-3 of Vilbert-Guesser State Estimator (Successful Case on Ground-Truth Dialogs)
}
\label{fig:guesser_example_3}
\end{figure*}

Where $|D|$ is the number of dialogues, and $M$ is the number of objects in each image. To view a concrete example of how object states are updated turn-by-turn, please refer to our supplementary materials (Figure \ref{fig:guesser_example_1}-\ref{fig:guesser_example_6}).

  \subsection{VilBERT-Questioner Model}
\begin{figure}[h]
\centerline{\includegraphics[width=0.7\linewidth]{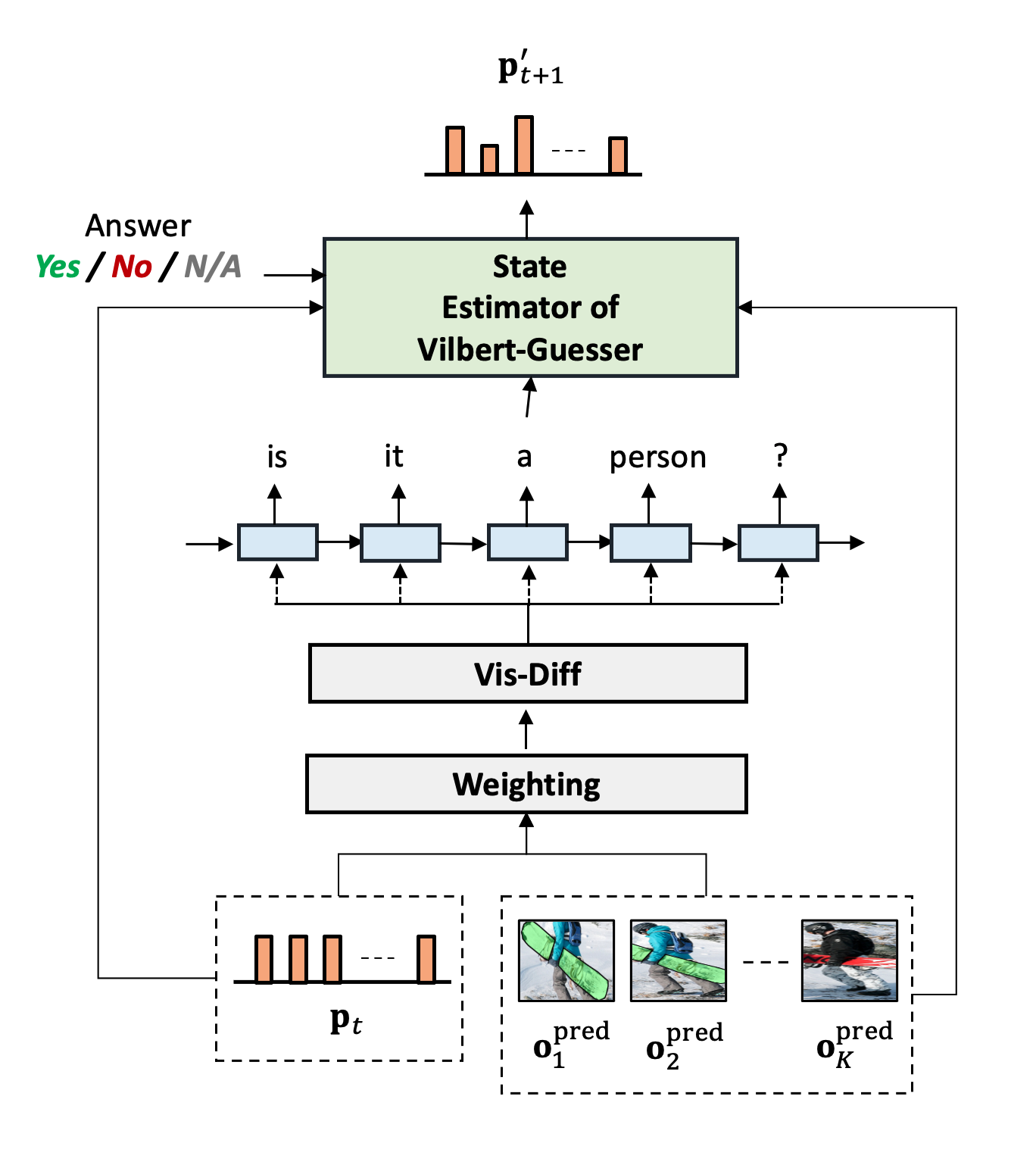}}

\caption{
Illustration of Vilbert-Questioner model.
}

\label{fig:qgen_vilbert}
\end{figure}

The Questioner model can be considered as a special case of referring expression generation problem. Given an image $I$ with $K$ regions/objects $\{\mathbf{o}^\text{pred}_1, ..., \mathbf{o}^\text{pred}_K\} $   and a dialog history $\{(q_1, a_1), ..., (q_{t-1}, a_{t-1})\} $, the Questioner model is expected to generate a new question $q_t$ that seeks useful information about the target object strategically. The model structure of VilBERT-Questioner is depicted in Figure-\ref{fig:qgen_vilbert}. 

The VilBERT-Questioner model is composed of a state-estimator, a state-reweighting layer, a vis-diff layer, and a question generator. In each turn, starting from uniformly distributed object states $p_{0}$ , we first re-weight visual features of all objects $\{\mathbf{o}^\text{pred}_1, ..., \mathbf{o}^\text{pred}_K\} $ with the last-turn object states $p_{t-1}$ . Then we feed the re-weighted object visual features into vis-diff module to derive the most distinctive feature of each object relative to others and merge the representation to $\mathbf{v}_t$ \cite{yu2016modeling,pang2020visual}. Then the language decoder (LSTM \cite{hochreiter1997long}) generates question conditioned on the encoder output $\mathbf{v}_t$.


The generated question together with its corresponding answer, is fed into the state-estimator, which is the pretrained VilBERT-Guesser, to get updated state belief $p_{t+1}$ as input to next-turn Vilbert-Questioner encoder.
The loss function of the VilBERT-Questioner model is defined as follows:
\begin{equation}
\label{eq:reweight}
L_{VilBERT-Questioner} = - \sum_{i=1}^{T}  \sum_{j=1}^{L}  \sum_{k=1}^{|V|}  y_{i,j,k} \cdot log (p_{i,j,k})
\end{equation}
Where $T$ is the number of turns in the dialog, $L$ is the length of questions, and $|V|$ is the size of vocabulary for the language decoder.

\section{Experiments}

\subsection{Dataset}
The \textit{GuessWhat?!} dataset \cite{de2017guesswhat} contains 155k dialogues with 821k question-answer pairs on 66k unique images with 134k unique objects. The answers are 52.2\%, 45.6\% and 2.2\% for (Yes,No,N/A) respectively. 84.6\% of the dialogues are successful games. We use the partitioned datasets with training (70\%), validation (15\%) and test (15\%) in all of our experiments, as specified in the original \textit{GuessWhat?!} dataset \cite{de2017guesswhat}.

\subsection{Evaluation Metrics}
\noindent\textbf{Independent Accuracy}
Independent accuracy refers to the percentage of correct predictions one agent achieves by isolating  it on the ground-truth data without interacting with other agents. The Oracle and Guesser models can be evaluated independently. 

\noindent\textbf{End-to-end Success Rate}. 
The Questioner models cannot be evaluated independently due to its dependency on dynamically generated dialog history. The Questioner models can only be evaluated jointly by having all three agents play the \textit{GuessWhat?!} game together and measuring the success rate at the end of the game, which is the percentage of games where the Guesser model makes correct guesses based on generated dialogues. 

\noindent\textbf{Semantic Diversity and Rate of Games with Repeated Questions}. One common problem for Questioner models is the generation of repeated questions within a game. Repeated questions reduces the opportunities to ask more meaningful questions. Therefore, we measure the percentage of games with at least one repeated question, as in previous work \cite{shekhar2018beyond}. 

\subsection{Experiment Settings}
The backbone of the VilBERT encoder \cite{lu2019vilbert} in all our models is adapted from the  official VilBERT implementation \footnote{\url{https://github.com/facebookresearch/vilbert-multi-task}}. We use Faster R-CNN model~\cite{ren2015faster} for feature extraction, pre-trained on Visual Genome dataset~\cite{krishna2017visual} with ResNet-101 backbone~\cite{he2016deep}. Relative coordinates and area are concatenated with feature vectors before feeding them into VilBERT. 
For the VilBERT-Oracle, the category embedding size is 512, and the number of bounding boxes $M$ is 100.
For the VilBERT-Guesser, the answer embedding size is 128 and the state accumulation coefficient $\alpha$ is 0.9.
For the VilBERT-Questioner decoder, the word embedding size is 512 and the number of  bounding boxes $K$ is 100.


\subsection{Results}

\subsubsection{The Oracle Model}
To the best of our knowledge, all existing work use the same baseline Oracle \cite{de2017guesswhat} except \cite{strub2018visual}. We compare the performance of the baseline oracles with the proposed VilBERT-Oracle. Further, we also modify the baseline Oracles by introducing image-level and object-level Faster-RCNN features as extra input for predicting answer.

From the results in Table-\ref{tab:oracle_independent}, we observe the following. First, introducing visual features increases the accuracy of baseline Oracle. This is intuitive since the baseline Oracle only relies on category/spatial information and not any visual information to predict any answer, it is prone to errors on challenging questions that need robust visual grounding. Second, VilBERT-Oracle further outperforms baseline Oracle, RCNN-Oracle and MultiHop Oracle. We attribute this to the two-way global/target fusion on top of VilBERT encoder, which not only supports matching of the target object with descriptive question, but also helps with contextually capturing the underlying relationships between objects in the image, therefore making it easier to answer more complex questions such as \textit{is it to the left of the women in red?}.

To corroborate to this observation, we further break down Oracle models' performances across different types of questions same as previous work \cite{shekhar2018beyond}, as in column 2 and 3 in Table-\ref{tab:oracle_by_type_question_dist}. From the table we see that VilBERT-Oracle indeed performs much better on all types of questions other than object type compared to baseline Oracle by 8\% - 19\%.

\begin{table}
\begin{center}
\begin{tabular}{|l|c|}
\hline
Oracle Models & Accuracy \\
\hline\hline
Baseline Oracle \cite{de2017guesswhat}& 78.5\% \\
\hline
RCNN-Oracle (Ours) & 81.7\% \\
Multi-hop FiLM Oracle\cite{strub2018visual}& 83.1\% \\
VilBERT-Oracle (Ours) & \textbf{85.0\% }\\

\hline
\end{tabular}
\end{center}
\caption{Comparison of Oracle Models (Independent Accuracy Evaluation)}
\label{tab:oracle_independent}
\end{table}

\subsubsection{The Guesser Model}
For Guesser models, we compare proposed VilBERT-Guesser with a comprehensive set of baselines and SOTAs under an independent evaluation setting. The input is an image along with the entire dialog history of question/answer pairs, and the output is the target object, all from ground-truth. 


From the results in Table-\ref{tab:guesser_independent}, we observe the following. First, Guesser models that utilize image features (VilBERT \cite{lu202012}, GST \cite{pang2020guessing}, ATT-R4 (w2v) \cite{deng2018visual}, and HACAN \cite{yang2019making}) achieve slightly higher accuracy compared to models that include no visual information (LSTM \cite{deng2018visual}, Guesser \cite{strubend}, and RIG \cite{shukla2019should}). Second, Guesser models that encode text at turn-level instead of dialog-level shows slightly better performance (GST\cite{pang2020guessing} and HACAN \cite{yang2019making}). Third, our VilBERT-Guesser outperforms all baseline models by an absolute margin of 10\%. Intuitively, the VilBERT-Guesser model encodes visual information of objects through pre-trained vision-linguistic layers of VilBERT; the turn-level state tracking and state accumulation mechanism, update the belief state with information from the VilBERT output, as it was used for referring expression comprehension. Both of these factors contribute to the improvement over state-of-the-art Guesser model. Please see supplementary material for example of object state update process.


\begin{table}
\begin{center}
\begin{tabular}{|l|c|}
\hline
Guesser Models & Accuracy \\
\hline\hline
Mask-RCNN (no gt bbox) \cite{bani2018adding} & 57.9\% \\
LSTM \cite{de2017guesswhat} & 61.3\% \\
PLAN \cite{zhuang2018parallel} & 63.4\% \\
Guesser \cite{strubend} & 63.8\% \\
RIG \cite{shukla2019should} & 64.2\% \\
12-in-1 Vilbert \cite{lu202012} & 65.7\% \\
GST \cite{pang2020guessing} & 65.7\%\\
ATT-R4 (w2v) \cite{deng2018visual} & 65.8\% \\
HACAN \cite{yang2019making} & 66.8\% \\
Multi-hop FiLM Guesser \cite{strub2018visual}& 69.5\% \\
\hline
Vilbert-Guesser (Ours) & \textbf{76.5\%} \\
\hline
\end{tabular}
\end{center}
\caption{Comparison of Guesser Models (Independent Accuracy Evaluation)}
\label{tab:guesser_independent}
\end{table}

\subsubsection{The Questioner Model}

 \textbf{End-to-end success rate}. Table-\ref{tab:qgen_e2e} compares different Questioner models in end-to-end self-play games based on success rate. The dialog sessions were generated by making Questioner and Oracle talking to each other, and having Guesser to make a final guess about the target object. Rows 1-5 correspond to baseline and SOTA Questioner model results, rows 6-10 refer to different combinations of proposed models, and rows 11-12 are two variants of the VilBERT-Questioner model. 

From the results (row 1-10), we observe the following. First, the state-of-the-art Questioner (VDST) only achieves slightly improvement over baseline Questioner (45.9\% over 44.6\%) when collaborating with baseline Oracle and Guesser. On the other hand, when combined with state-of-the-art Guesser (GST), VDST achieves much higher performance (50.6\%). Our speculation is that GST and VDST share very similar model structures, which promotes mutual understanding in self-play games \cite{pang2020visual,pang2020guessing}. Similarly for GDSE-SL, the Guesser and Questioner share the same encoder that encodes visually grounded dialog states. This might promote mutual understanding of the two models in end-to-end games. Second, our VilBERT-Questioner, combined with baseline Oracle and Guesser (row-8), also achieves higher end-to-end success rate over VDST (52.5\% over 45.9\%). Third, for row 9 and 10, when we introduce two or three proposed models into the game, the performance continuously improves (55.7\% and 62.8\% respectively). As contrast, as in row 6 and 7, when only VilBERT-Oracle or VilBERT-Guesser is introduced in the game, the improvement is minimal. We will discuss details in Ablation Study.



Please note that we did not include RL-based models for comparison of success rate, since this work focuses on supervised learning. Previous work also show evidence that RL-based Questioner models may generate unnatural questions as indicated by skewed distribution across different visual attributes \cite{mazuecos2020role,shekhar2018beyond}. As can be seen in Table-\ref{tab:qgen_e2e}, RL \cite{strubend} and VDST \cite{pang2020visual}, which are both RL-based models tend to generate more questions about object and location, while BL \cite{de2017guesswhat}, SL \cite{shekhar2018beyond}, CL\cite{shekhar2018beyond}, and our VilBERT-Questioner have less skewed distributions across different question types. 

\textbf{Rate of Games with Repeated Questions}. One common issue for Questioner models is question repetition \cite{shekhar2018beyond}. We compare the rate of games with repeated questions, for the baseline Questioner models and our VilBERT-Questioner. The results are reported in Table-\ref{tab:repeated_q}. From the results we can see that our VilBERT-Questioner has the lowest rate of games with repeated questions.


\begin{table*}[t]
\begin{center}
\begin{tabular}{|llllc|}
\hline
&Oracle & Guesser & Questioner & Success Rate \\
\hline\hline
1 & Baseline Oracle \cite{de2017guesswhat} & Baseline Guesser \cite{strubend} & Baseline Questioner \cite{strubend}& 44.6\% \\
2 & Baseline Oracle \cite{de2017guesswhat} & Baseline Guesser \cite{strubend} & VDST \cite{pang2020visual}  & 45.9\% \\
3 & Baseline Oracle \cite{de2017guesswhat} & GDSE-SL  \cite{shekhar2018beyond} & GDSE-SL \cite{shekhar2018beyond} & 47.8\% \\
4 & Baseline Oracle \cite{de2017guesswhat} & Guesser (MN)  \cite{zhao2018improving} & TPG \cite{zhao2018improving} & 48.8\% \\
5 & Baseline Oracle \cite{de2017guesswhat} & GST \cite{pang2020guessing} & VDST \cite{pang2020visual} & 50.6\% \\
\hline
\hline\hline
6 & Baseline Oracle \cite{de2017guesswhat} & \textbf{ Vilbert-Guesser (Ours)} & VDST \cite{pang2020visual}  & 47.5\% \\
7 & \textbf{Vilbert-Oracle (Ours) } & Baseline Guesser \cite{strubend} & VDST \cite{pang2020visual}  & 47.8\% \\
8 & Baseline Oracle \cite{de2017guesswhat} & Baseline Guesser \cite{strubend} & \textbf{VilBert-Questioner 
(Ours)} & 52.5\% \\

9 & \textbf{Vilbert-Oracle (Ours)} & \textbf{Vilbert-Guesser (Ours)} & VDST \cite{pang2020visual}  & 55.7\%\\
10 & \textbf{Vilbert-Oracle (Ours)} & \textbf{Vilbert-Guesser (Ours)} & \textbf{Vilbert-Questioner (Ours)} & \textbf{62.8\%} \\
\hline\hline
11 & \textbf{Vilbert-Oracle (Ours)} & \textbf{Vilbert-Guesser (Ours)} & \textbf{Vilbert-Questioner (w/ fine-tune)} & 57.0\% \\
12 & \textbf{Vilbert-Oracle (Ours)} & \textbf{Vilbert-Guesser (Ours)} & \textbf{Vilbert-Questioner (w/o fine-tune)} & \textbf{62.8\%} \\

\hline
\end{tabular}
\end{center}
\caption{Comparison of Different Questioner Models in End-to-End Evaluation}
\label{tab:qgen_e2e}
\end{table*}

\begin{table*}[t]
\begin{center}
\begin{tabular}{|l|cc|cccccc|c|}
\hline
Type & \makecell{Baseline \\ Oracle  \cite{de2017guesswhat}} & \makecell{VilBERT-\\ Oracle} & BL\cite{de2017guesswhat} & SL \cite{shekhar2018beyond} & CL \cite{shekhar2018beyond} & RL \cite{strubend}&  VDST \cite{pang2020visual}& \makecell{VilBERT-\\Questioner} &  Human \\
\hline\hline
Object & 94\% & 94\% & 49.00 & 48.08 &  46.40 & 24.00 & 36.44 & 65.23 & 38.12\\
Color & 63\% & \textbf{82\%} & 2.75 & 13.00 &  12.51 & 0.12 & 0.01 & 9.1 & 15.50\\
Shape & 67\% & \textbf{75\%} & 0.00 & 0.01 &  0.02 & 0.00 & 0.00 & 0.00 & 0.30\\
Size & 60\% & \textbf{77\%} & 0.02 & 0.33 &  0.39 & 0.02 & 0.01 & 0.01 & 1.38\\
Texture & 70\% & \textbf{83\%} & 0.00 & 0.33 &  0.15 & 0.01 & 0.00 & 0.00 & 0.89 \\
Location & 67\% & \textbf{77\%} & 47.25 & 37.09 &  38.54 & 74.80 & 64.80 & 25.60 & 40.00 \\
Action & 65\% & \textbf{81\%} & 1.34 & 7.97 &  7.60 & 0.66 & 0.30 & 5.04 &7.59 \\
Other & 75\% & \textbf{82\%} & 1.12 & 5.28 & 5.90 & 0.49 & 0.03 & 1.95 & 8.60 \\

\hline
\end{tabular}
\end{center}
\caption{Oracle Accuracy by Types of Question and Question Distribution for Models. }
\label{tab:oracle_by_type_question_dist}
\end{table*}

\begin{table}
\begin{center}
\begin{tabular}{|l|c|}
\hline
Questioners &  \makecell{\% Games with  Repeated Q's } \\
\hline\hline
GDSE-BL \cite{de2017guesswhat} &  93.50 \\
GDSE-SL \cite{shekhar2018beyond} &  55.80 \\
CL \cite{shekhar2018beyond} & 52.19 \\
RL \cite{strubend} &  96.47 \\
VDST \cite{pang2020visual} &  40.05\\
VilBERT-Questioner &  \textbf{32.56} \\
\hline
Human &  N/A \\
\hline
\end{tabular}
\end{center}
\caption{Rate of Games with Repeated Questions of Different Questioner Models. Note: * the VDST Model Used Here is the Model Trained in Supervised Learning Setting.}
\label{tab:repeated_q}
\end{table}

\section{Ablation Study}

\subsection{Variants of Guesser Model}
In this paper, we decompose the dialog history to turns, and feed question representation to the VilBERT-Guesser encoder, instead of concatenating the entire dialogue history together \cite{deng2018visual,strubend,shukla2019should,lu202012}. We compare two methods, namely injecting answer information to the post-layers of the state-estimator (\textbf{Post-Fusion}), versus concatenating a pair of question and answer as input to the model (\textbf{Pre-Concatenation}).

From Table-\ref{tab:injecting_answers} we observe that both variants outperforms the previous work \cite{lu202012} significantly by 9\%-11\%. Second, the Post-Fusion method further outperforms Pre-Concatenation method by 2\%. We hypothesize that pre-trained vision-linguistic models like VilBERT may be best used to its strength if we isolate the input to a way similar to how the model has been pretrained, in this case, on single-turn descriptive questions. 

\begin{table}
\begin{center}
\begin{tabular}{|l|c|}
\hline
Models & Accuracy \\
\hline\hline
12-in-1 VilBERT: Concatenate Entire Dialog & 65.7\% \\
\hline
Answer Pre-Concatenation (Ours) & 74.3\% \\
Answer Post-Fusion (Ours) & \textbf{76.5\%} \\

\hline
\end{tabular}
\end{center}
\caption{Different Variants of Answer Fusion}
\label{tab:injecting_answers}
\end{table}

\subsection{Variants of Questioner Model}
For Vilbert-Questioner, the weights of state-estimator is loaded from pretrained VilBERT-Guesser. We compare two variants: fine-tuning the state-estimator together with question generator (\textbf{w/ fine-tune}), versus freezing the weights of state-estimator while training the rest of the Vilbert-Questioner (\textbf{w/o fine-tune}). These results are reported in row 11-12 of Table-\ref{tab:qgen_e2e}. The results show that the VilBERT-Questioner \textbf{w/ fine-tune} performs poorer than the VilBERT-Questioner \textbf{w/o fine-tune}. This is intuitive since the state-estimator learned by VilBERT-Guesser is already good at inferring object states turn-by-turn, fine-tuning it with the question generation objective may confuse the state-estimator to go astray from predicting the correct states and thus make the end-to-end performance worse. 
\subsection{Variants of Combinations of the Three Models}
As is shown in the ablation study (Table-\ref{tab:qgen_e2e} row 6,7 and 9), when only the VilBERT-Oracle or VilBERT-Guesser is introduced to the end-to-end game, the overall accuracy has shown minor improvement over the baselines. Whereas when both are introduced, the end-to-end accuracy has improved significantly (55.7\% over 50.6\%). We investigate the possible causes as follows. 

\begin{figure}[t]
\begin{center}
\includegraphics[width=1.0\linewidth]{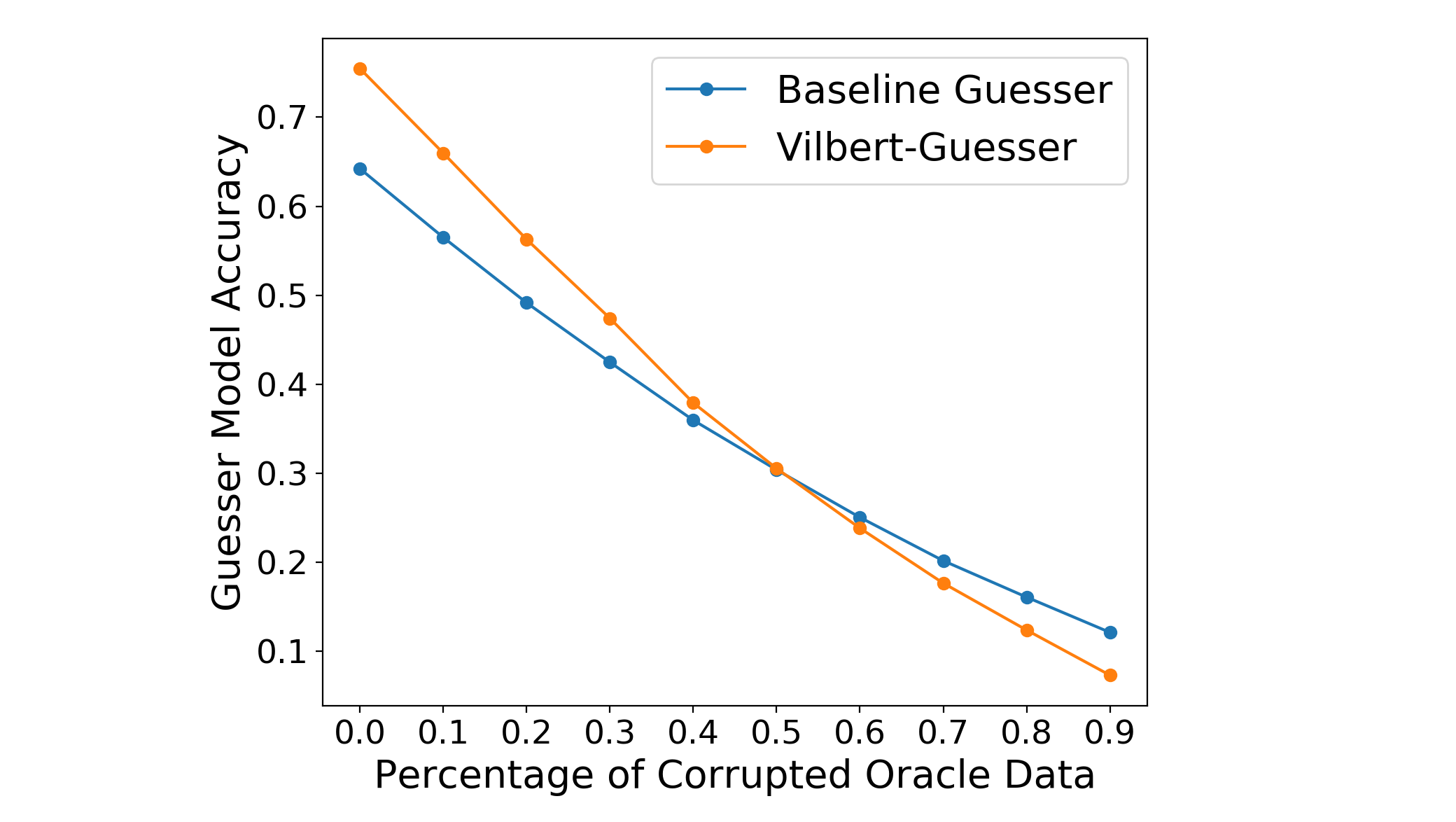}
\end{center}
   \caption{Performance of Two Guesser Models on Corrupted Oracle Data with Varying Corruption Ratios}
\label{fig:guesser_on_corrupted_oracle}
\end{figure}

When Oracle model's performance is very poor, even a highly accurate Guesser may not achieve a high end-to-end success rate. To simulate this effect, we keep a fixed ground-truth dialog and add random errors to the answer data with varying levels (10\%-90\%) by toggling the yes/no answer. We run both the baseline guesser and our VilBERT-Guesser on this corrupted dataset to study model performance deterioration. The results are reported in Figure \ref{fig:guesser_on_corrupted_oracle}. When the corruption ratio is low (10\%-40\%), VilBERT-Guesser consistently outperforms the baseline Guesser. When the Oracle data is corrupted to a large ratio ($\geq$ 50\%), the accuracy of VilBERT-Guesser drops faster than baseline model, suggesting that it is more sensitive to the correctness of the Oracle output. This partially explains why row-6 shows little improvement over row-2 in Table-\ref{tab:qgen_e2e}, since in both settings the baseline Oracle is expected to have lower accuracy.

\begin{table}
\begin{center}
\begin{tabular}{|p{1.2 cm}|p{1.2 cm}|p{1.1cm}| p{1.1cm} | p{1cm} |}
\hline
\multicolumn{2}{|c|}{\multirow{2}{*}{}}&  \multicolumn{3}{c|}{VilBERT-Guesser}\\
\cline{3-5}
\multicolumn{2}{|c|}{} &  Correct &  Wrong & Total\\
\hline\hline
\multirow{3}{1.2cm}{Baseline Guesser} & Correct & 7565 & 1993 & 47.8\% \\
& Wrong & 3573 & 6864 & 52.2\% \\
& Total & 55.7\% & 44.3\% &  \\
\hline
\end{tabular}
\end{center}
\caption{Confusion Matrix of End-to-End Dialogue Success Rate Generated from Baseline Guesser and VilBERT-Guesser Using VilBERT-Oracle. }
\label{tab:confusion_matrix_vilbert_guesser}
\end{table}

Second, only introducing a better Oracle, while keeping a Guesser that is less sensitive to the correct/wrong answers, may also not help end-to-end success rate, as indicated in row-7 versus row-2 in Table-\ref{tab:qgen_e2e}. To demonstrate this, we compute a confusion matrix (Table-\ref{tab:confusion_matrix_vilbert_guesser}) between baseline Guesser and VilBERT-Guesser, both run on dialogs generated with VilBERT-Oracle and VDST \cite{pang2020visual}. From the table, it is clear that given a better Oracle (VilBERT-Oracle), VilBERT-Guesser is able to make more correct guess compared to a baseline Guesser. For more details of baseline Guesser and VilBERT-Guesser model behavior, please refer to examples in our supplementary materials.

Examining row 6-10 together, we argue that sharing a similar visual-linguistic encoder across the three agents may be beneficial for the end-to-end game. 

\section{Conclusion}
In this paper, we propose three novel models, VilBERT-Oracle, VilBERT-Questioner and VilBERT Guesser for the \textit{GuessWhat?!} game. 
The proposed models take advantage of a pretrained visual-linguistic encoder (VilBERT\cite{lu2019vilbert}) that has shown state-of-the-art performance in multiple vision-language tasks especially in VQA and referring expression comprehension. A state-estimator is introduced to the Guesser and Questioner model to handle object state update turn-by-turn. Experimental results show that feeding the VilBERT model with turn-level text description is better than feeding a long dialog history, in accordance with how the VilBERT model has been pretrained. Ablation study suggests that a shared vision-linguistic encoder may be beneficial for such three-agent games. For future work, we plan to explore different reinforcement learning approaches with Questioner and Guesser to further improve end-to-end success rates. 
{\small
\bibliographystyle{ieee_fullname}
\bibliography{egbib}
}
\clearpage
\section*{Supplementary Material}
In this section, we present supplementary experiemnts and results.
\section{Details of Experiment Setting}
For our Guesser model, the object states are initialized uniformly on the candidate objects in the images provided in the \textit{GuessWhat?!} dataset. We use AdamW optimizer \cite{loshchilov2018fixing} with a learning rate of 1e-5. We usually observe convergence of the Guesser model after 3-4 epochs. For our Oracle model, we use AdamW optimizer with a learning rate of 1e-5. We usually observe convergence of the Oracle model after 3 epochs. For our Questioner model, we use AdamW optimizer with a learning rate of 1e-3. We also keep the state-estimator freezed during training.

\section{Ablation Studies}
\subsection{Oracle Versus Guesser}
In Table-\ref{tab:guesser_behavior_example}, we present examples where the following four combinations of Oracle and Guesser Model model are compared on the same test data: (1) Baseline Oracle + Baseline Guesser; (2) VilBERT Oracle + Baseline Guesser; (3) Baseline Oracle + VilBERT Guesser; (4) VilBERT Oracle + VilBERT Guesser. From the examples we can see that both Oracle and Guesser contribute to the final correct guessing. Neither a stronger Oracle nor a stronger Guesser alone can achieve a superior end-to-end performance, with the other two models fixed on baseline models. Only when both models are stronger and also understand each other, the final guessing is more likely to be correct.
\subsection{A Weak Oracle}
We also experimented with the setting of [baseline Oracle, VilBERT Guesser, VilBERT Questioner], and the end-to-end success rate dropped from 62.8\% to 52.7\%.
\subsection{Using Advanced Language Encoder in Baseline Oracle Model}
We experimented with the baseline Oracle model that uses the same object encoder (Faster-RCNN) \cite{ren2015faster} with a pre-trained BERT model \cite{devlin2018bert} as language encoder. The final accuracy is 78.67\%, which is lower than our proposed model (85.0\%).
\section{Reinforcement Learning Training}
Although the present paper focuses on supervised learning on the \textit{GuessWhat?!} task, we also experimented with reinforcement learning training of the agents. More specifically, we first train the three models, namely Guesser, Questioner and Oracle in supervised setting as described previously. Then we update the Questioner model with policy gradient using the \textbf{REINFORCE} algorithm \cite{williams1992simple}. After 3 epochs, the final end-game success rate on the same test set is 0.65. We haven't fully fine-tuned our Questioner model in this setting, due to the significant computation time to generate questions and update states with our VilBERT-based models and the vast number of sampling epochs required in reinforcement learning. We consider this as one of our major future work to propose more efficient sampling algorithm and and speed up the reinforcement learning training process.

\section{Language Diversity of Generated Questions}
To evaluate language diversity of generated questions, we utilize \textbf{Self-Bleu} \cite{zhu2018texygen} to compare our VilBERT-Questioner and the state-of-the-art Questioner (VDST). The results are shown in Table \ref{tab:selfbleu}. We can see that our Questioner model has higher language diversity compared to SOTA Questioner after supervised learning.

\begin{table}
\begin{center}
\begin{tabular}{|l|c|c|c|}
\hline
Models & 2-Gram & 3-Gram & 4-Gram \\
\hline\hline
VDST & 0.63 & 0.55 & 0.47 \\
Our Questioner & \textbf{0.54} & \textbf{0.44} & \textbf{0.34} \\
\hline

\hline
\end{tabular}
\end{center}
\caption{Self-Bleu for Language Diversity of Questioner Models (Lower is better). }
\label{tab:selfbleu}
\end{table}

\section{Successful Cases}
\begin{figure*}[t]
\centerline{\includegraphics[width=\linewidth]{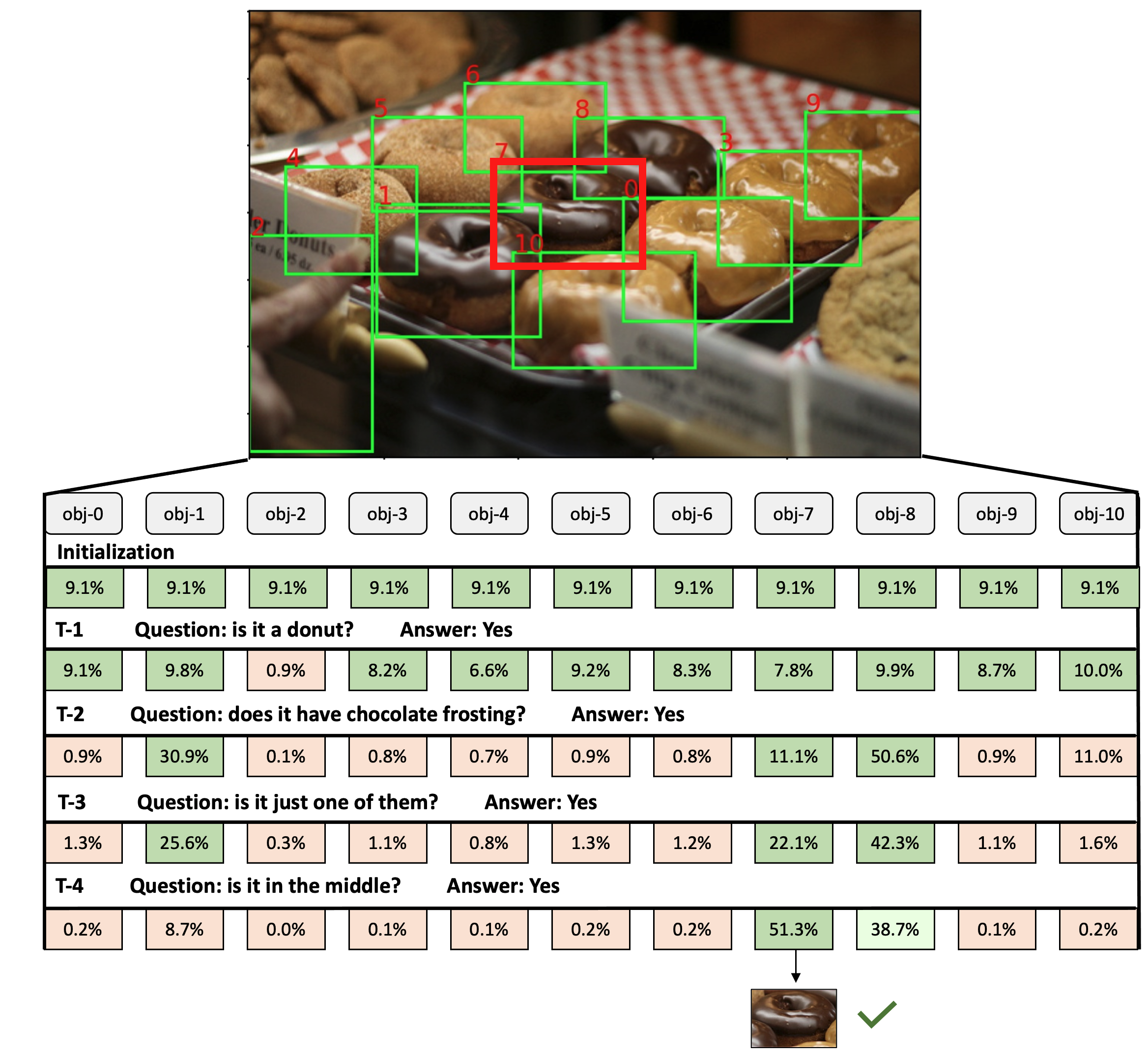}}
\caption{
Example-1 of Vilbert-Guesser State Estimator (Successful Case on Ground-Truth Dialogs)
}
\label{fig:guesser_example_1}
\end{figure*}

\begin{figure*}[t]
\centerline{\includegraphics[width=\linewidth]{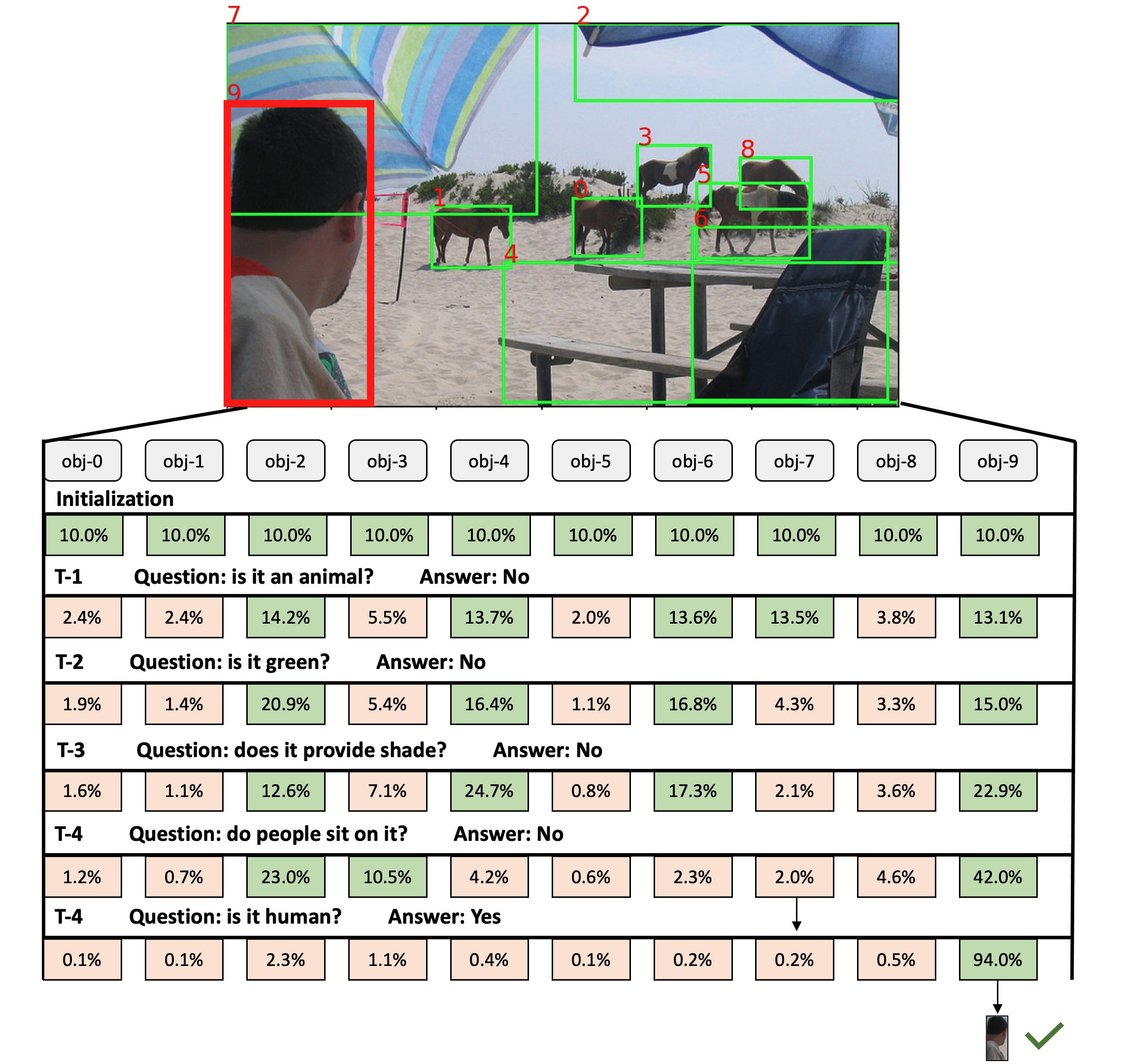}}

\caption{
Example-4 of Vilbert-Guesser State Estimator (Successful Case on Ground-Truth Dialogs)
}

\label{fig:guesser_example_4}
\end{figure*}

Figure-\ref{fig:guesser_example_1} - Figure-\ref{fig:guesser_example_4} present four successful cases of proposed models. We show how object state beliefs are updated by our VilBERT Guesser model from turn to turn. You can see that the state estimator eliminates the unlikely objects according to the question and answer in each turn, and narrows down to the final target object. The accurate estimation of object states turn-by-turn can be largely attributed to the pretrained VilBERT encoder, which already shows state-of-the-art performance on referring expression comprehension tasks.

\begin{table*}
\centering
\begin{tabular}{|p{7cm}|p{7cm}|}
\hline

\makecell[l]{
\raisebox{-\totalheight}{\includegraphics[width=0.4\textwidth]{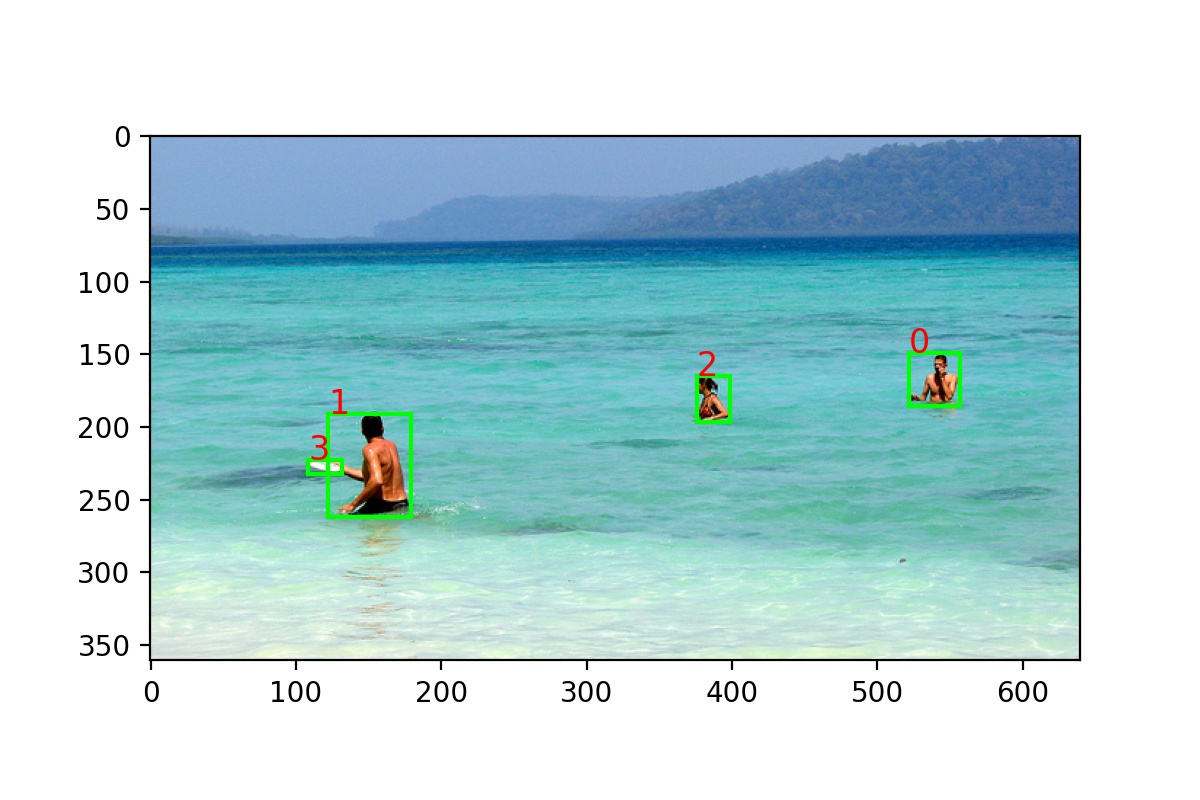}} \\
\textbf{Ground-truth object: 2} \\
\\
\textbf{Baseline Oracle + Baseline Guesser: 0} \\
is it a person?|yes \\ 
is it the person on the left?|no \\ 
is it the person on the right?|yes \\ \\
\textbf{Vilbert Oracle + Baseline Guesser:1} \\
is it a person?|yes \\ 
is it the person on the left?|no \\ 
is it the person on the right?|no \\ \\
\textbf{Baseline Oracle + Vilbert Guesser: 0} \\
is it a person?|yes \\ 
is it the person on the left?|no \\ 
is it the person on the right?|yes \\ \\
\textbf{Vilbert Oracle + Vilbert Guesser:2} \\
is it a person?|yes \\ 
is it the person on the left?|no \\ 
is it the person on the right?|no \\ \\
}  &

\makecell[l]{
\raisebox{-\totalheight}{\includegraphics[width=0.4\textwidth]{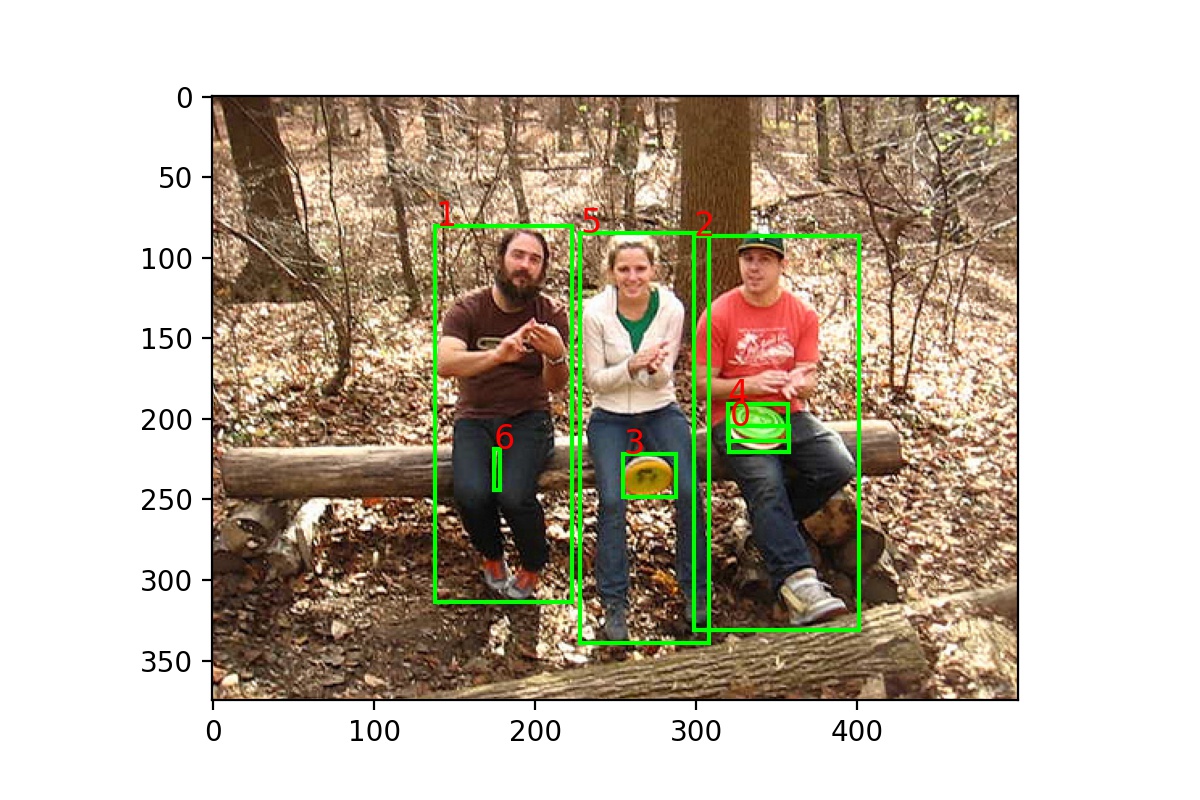}} \\
\textbf{Ground-truth object: 2} \\
\\
\textbf{Baseline Oracle + Baseline Guesser: 1} \\
is it a person?|yes \\ 
is it a female?|yes \\ 
is she wearing a hat?|yes \\ \\
\textbf{Vilbert Oracle + Baseline Guesser:1} \\
is it a person?|yes \\ 
is it a female?|no \\ 
is it a man?|yes \\
is he wearing a hat?|yes \\ \\
\textbf{Baseline Oracle + Vilbert Guesser: 5} \\
is it a person?|yes \\ 
is it a female?|yes \\ 
is she wearing a hat?|yes \\ \\
\textbf{Vilbert Oracle + Vilbert Guesser:2} \\
is it a person?|yes \\ 
is it a female?|no \\ 
is it a man?|yes \\
is he wearing a hat?|yes \\ \\
}  \\

\hline
\end{tabular}
\caption{Examples of Guesser Models Behavior Conditioned on Oracle Models}
\label{tab:guesser_behavior_example}
\end{table*}

\section{Failure Cases}

\begin{figure*}[t]
\centerline{\includegraphics[width=\linewidth]{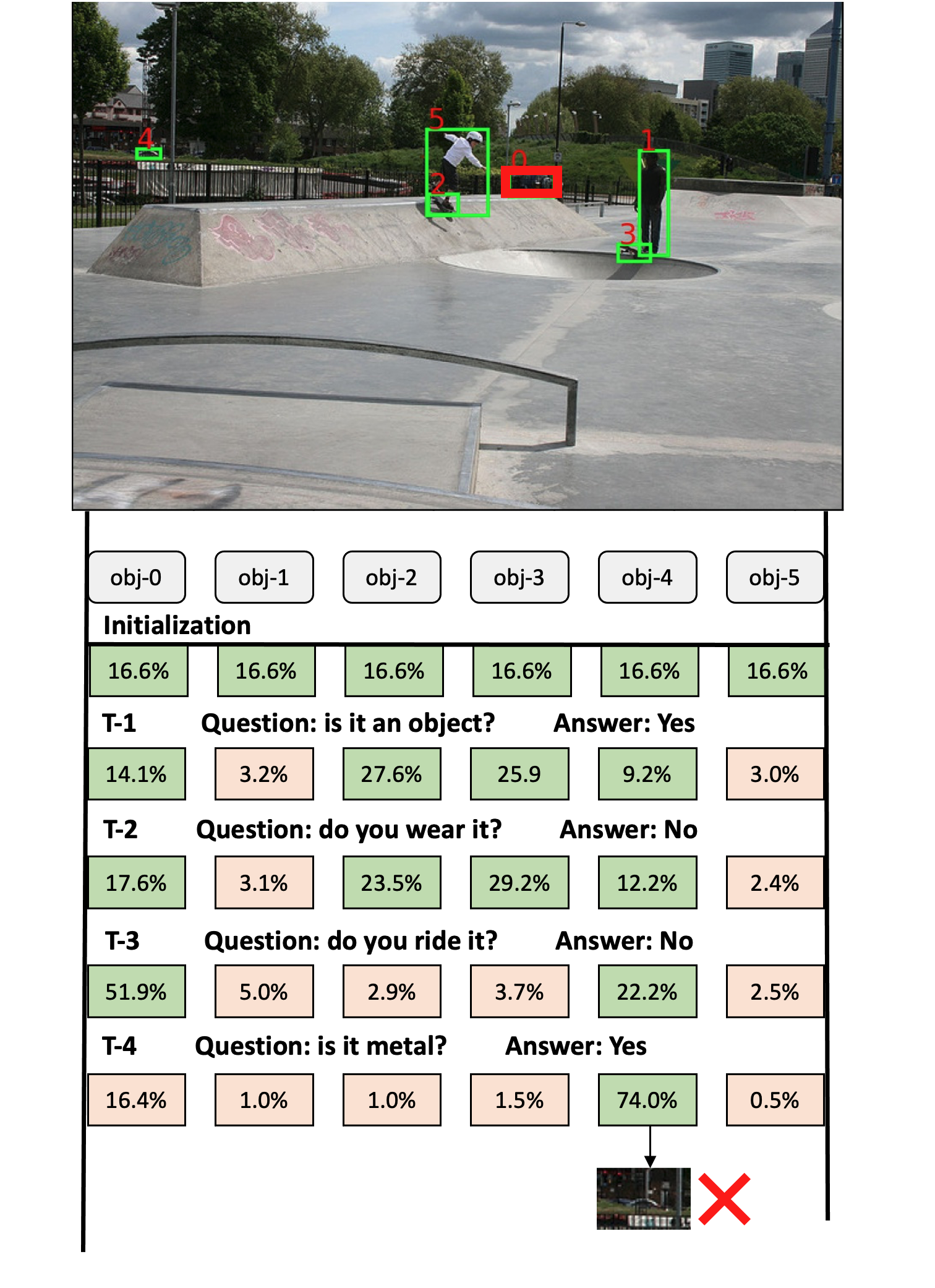}}
\caption{
Example-1 of Vilbert-Guesser State Estimator (Failure Case on Ground-Truth Dialogs)
}
\label{fig:guesser_example_5}
\end{figure*}

\begin{figure*}[t]
\centerline{\includegraphics[width=\linewidth]{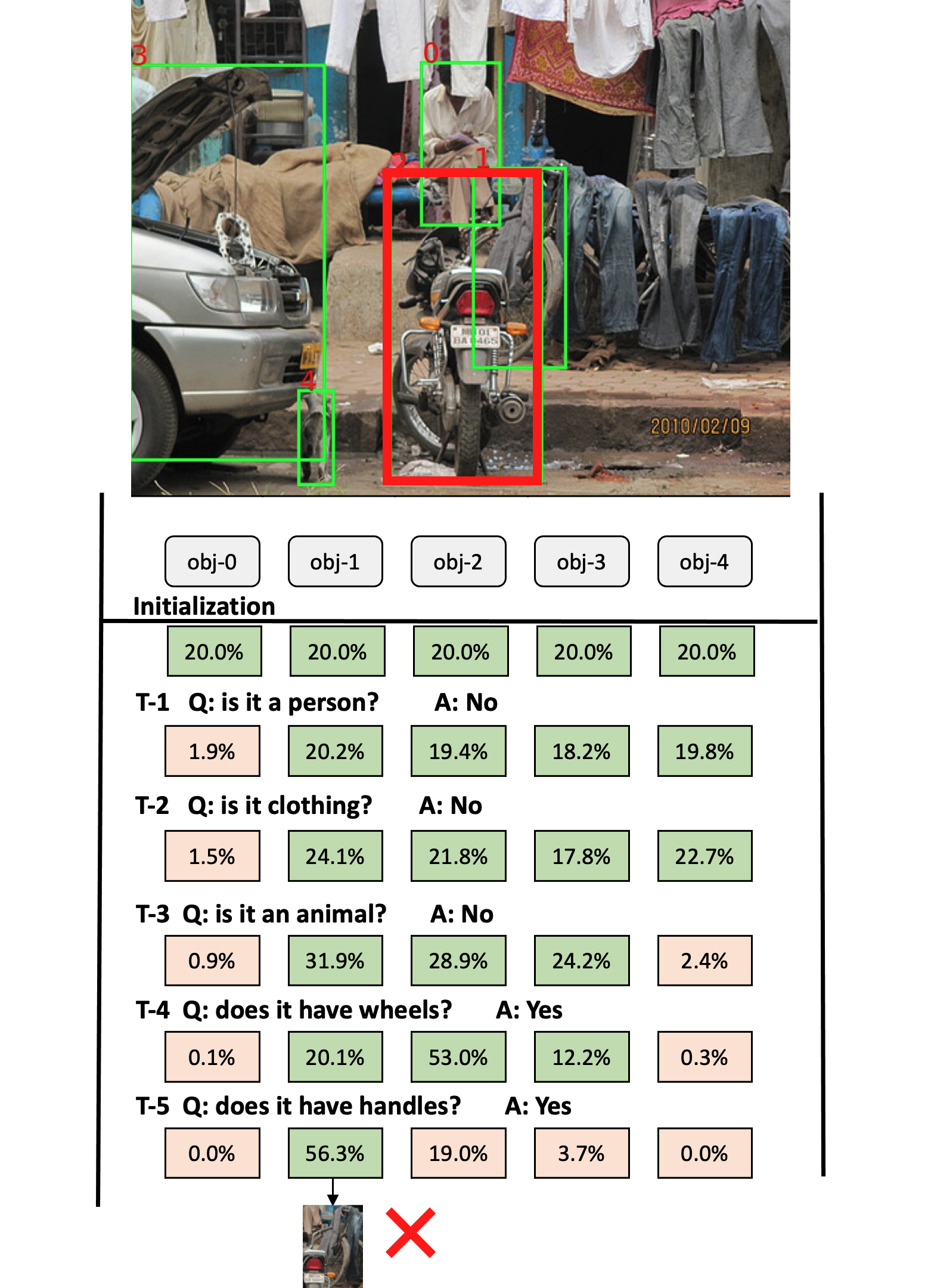}}

\caption{
Example-2 of Vilbert-Guesser State Estimator (Failure Case on Ground-Truth Dialogs)
}

\label{fig:guesser_example_6}
\end{figure*}

Figure-\ref{fig:guesser_example_5} - Figure-\ref{fig:guesser_example_6} present two failure cases of proposed Guesser model. For Figure-\ref{fig:guesser_example_5}, we can see that the ground-truth dialog (between two humans) ends prematurely, in which both object-0 and object-4 (vehicles) are made of metal. The Guesser model has made a wrong guess based on the premature dialog to predict the target as object-4 while ground-truth target is object-0. In Figure-\ref{fig:guesser_example_6}, similar pattern can be observed. Both object-1 and object-2 satisfies the conditions: not a person; not clothing; not animal; has wheels and has handles.

\end{document}